\pdfoutput=1
\documentclass[11pt]{article}

\usepackage{coling}

\usepackage{times}
\usepackage{latexsym}

\usepackage[T1]{fontenc}
\usepackage[utf8]{inputenc}

\usepackage{microtype}

\usepackage{inconsolata}

\usepackage{graphicx}
\usepackage{booktabs}
\usepackage{amsmath}
\usepackage{amssymb}
\usepackage{amsfonts}
\usepackage{enumitem}
\usepackage{multirow}
\usepackage{tikz}
\usepackage{pgfplots}
\pgfplotsset{compat=1.5.1}
\def\addlegendimage{\csname pgfplots@addlegendimage\endcsname}
\usetikzlibrary{patterns}
\usetikzlibrary{pgfplots.groupplots}
\usetikzlibrary{
  pgfplots.colorbrewer,
}

\definecolor{purple}{rgb}{0.56, 0.0, 1.0}

\newcommand\blfootnote[1]{%
  \begingroup
  \renewcommand\thefootnote{}\footnote{#1}%
  \addtocounter{footnote}{-1}%
  \endgroup
}

\usepackage{inconsolata}

\usepackage{mathtools}

\usepackage{amsmath, amssymb, amsthm}
\usepackage{enumerate}
\usepackage{amsfonts} 
\usepackage{lipsum} 
\usepackage{algorithm}
\usepackage{algorithmic}

\usepackage{graphicx}
\usepackage{transparent}
\usepackage{soul}
\usepackage{calc}
\definecolor{darkred}{HTML}{bb0000}

\usepackage{caption} 
\usepackage{courier}

\usepackage{booktabs}
\usepackage{multirow}
\usepackage{comment}
\usepackage{makecell}
\usepackage{arydshln}
\usepackage{color,framed}
\usepackage{xcolor,colortbl}

\usepackage{subcaption}
\usepackage{gensymb}

\usepackage{enumitem}

\usepackage{nccmath}
\usepackage{hyperref}
\usepackage{grffile}
\usepackage{tabularx}
\usepackage{nicefrac}       % compact symbols for 1/2, etc.
\usepackage{adjustbox}

\usepackage{wrapfig}
\usepackage{hhline}
\usepackage{tikz}

\title{CharacterGPT: A Persona Reconstruction Framework for Role-Playing Agents}

\author{Jeiyoon Park$^{1}$,\;\;\; Chanjun Park$^{2\dagger}$, \;\;\; Heuiseok Lim$^{2\dagger}$
\\
  $^{1}$ SOOP,\;\; $^{2}$ Korea University  \\
  \texttt{naruto@sooplive.com} \\
  \texttt{\{bcj1210, limhseok\}@korea.ac.kr} \\
}

\begin{document}
\maketitle
\begin{abstract}
\blfootnote{$^\dagger$ Corresponding Author}
The recent introduction of the Assistants API highlights its potential for large language models (LLMs) in role-playing agents (RPA). However, maintaining consistent character personas remains a significant challenge due to variability in information extraction, which frequently omits critical elements such as backstory or interpersonal relationships. To address this limitation, we introduce CharacterGPT, a framework designed to dynamically reconstruct character personas through Character Persona Training (CPT). This approach incrementally updates personas by extracting traits from chapter-wise novel summaries, reflecting the progression of the narrative. Our framework is evaluated through Big Five personality evaluations and creative tasks, in which characters generate original narratives, demonstrating the efficacy of CharacterGPT in preserving persona consistency. The code and results are available at \url{https://github.com/Jeiyoon/charactergpt}
% \footnote{The code was released following anonymization for review purposes.}
\end{abstract}

\section{Introduction}

\begin{figure}[t!]
    \centering 
    \includegraphics[width=0.9\linewidth]{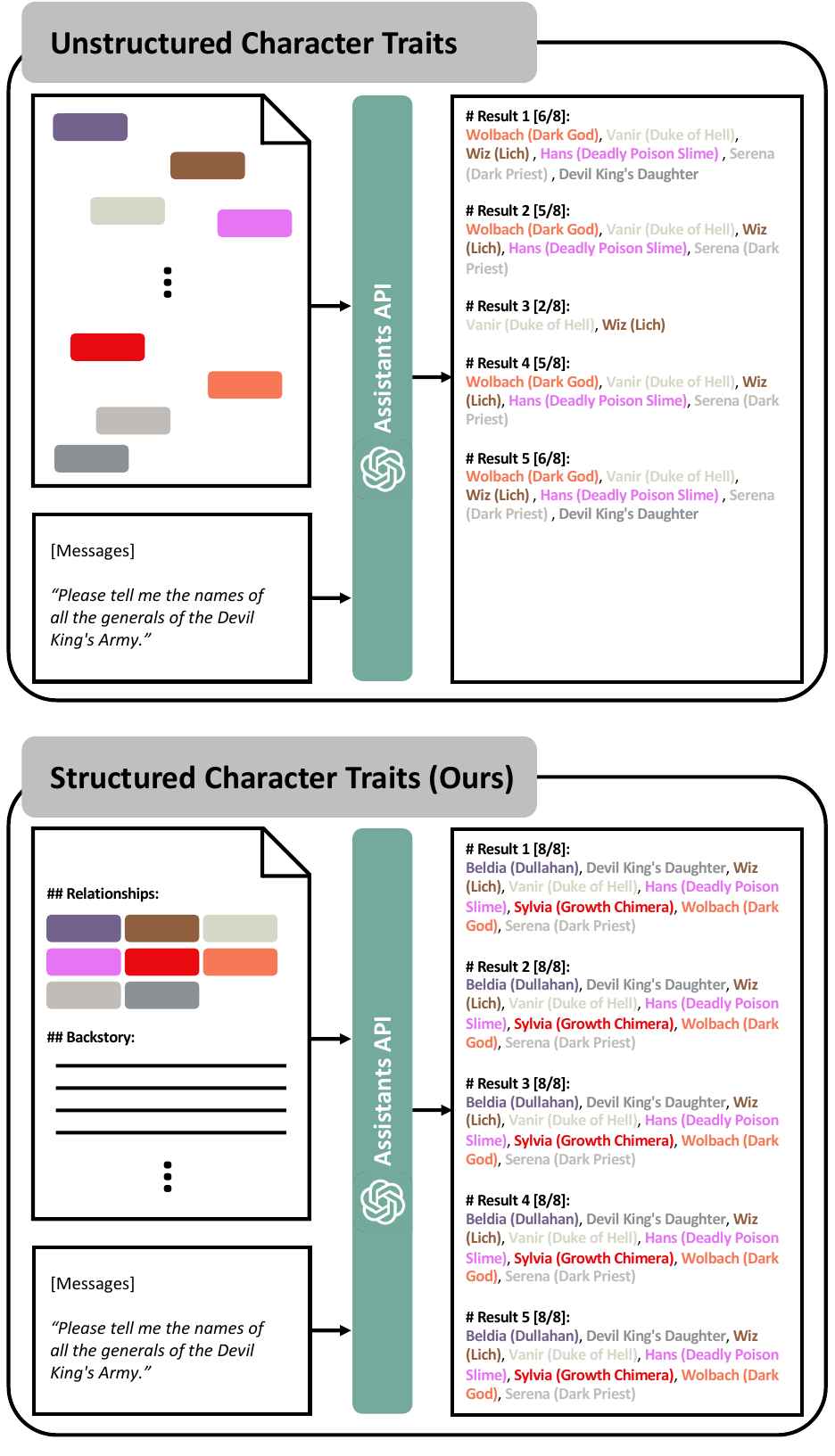}
    \caption{Comparison of response accuracy between persona-based GPT-4 assistants utilizing unstructured versus structured character traits as input. When provided with unstructured traits, the assistant demonstrates limited success in generating accurate responses. In contrast, the use of structured traits significantly improves the correctness of the assistant’s responses.}
    \label{fig:intro}
\end{figure}

The rapid advancements in large language models (LLMs) have positioned them as the core module of various AI systems \cite{openai2023gpt4,OpenAI_ChatGPT,claude,gemini1_5,deepseekai2024deepseekv3technicalreport,deepseekai2025deepseekr1incentivizingreasoningcapability}, enabling a wide range of applications. Building on this progress, the recent introduction of the Assistants API \cite{openai2023assistants}, a tool designed for document-based information retrieval, demonstrates the potential of LLM in multiple domains, especially in role-playing agents (RPA) \cite{kim2019understanding,yu-etal-2023-personality,jiang2023personallm,park2023generative,wang2023rolellm,zhang2024revealingchallengedetectingcharacter,kong-etal-2024-better,wang2025opencharactertrainingcustomizableroleplaying}. However, RPAs that rely solely on documents as input often face problems of inconsistent information extraction, where key personality traits or background knowledge are omitted, leading to degraded persona coherence \cite{sadeq-etal-2024-mitigating}. For example, as illustrated in Figure \ref{fig:intro}, when the Assistants API is provided with an unstructured Wiki document about the novel \textit{God's Blessing on This Wonderful World!}, it often fails to provide accurate responses, while structured character traits produce more reliable, role-specific answers.

In this paper, we propose a novel framework called \textit{CharacterGPT}, which addresses this challenge through a structured persona reconstruction process. Drawing inspiration from cognitive memory models, we introduce \textit{Character Persona Training (CPT)}, a method that incrementally updates character personas by extracting traits from chapter-wise summaries of novels. This approach mirrors how human memory consolidates information into schemas over time \cite{van2020optimize}, enabling more consistent and contextually appropriate responses from RPAs.

\textit{CPT} operates by identifying eight essential traits—\textit{personality, physical description, motivations, backstory, emotions, relationships, growth and change}, and \textit{conflict}—based on character analysis literature \cite{forster1927aspects,reams2015characterization}. For each chapter of a novel, these traits are extracted from summaries and appended to a character’s evolving persona, forming a document that reflects the character’s development in chronological order. Note that the extracted traits are updated separately to ensure they remain distinct and are not coalesced. This reconstructed persona document is then used as input to the Assistants API, allowing it to generate more contextually accurate and coherent responses based on the character’s evolving identity. This framework minimizes information loss and computational cost associated with traditional document-based retrieval methods, as it systematically organizes and updates persona traits over time. Moreover, by generating personas at different narrative points, \textit{CharacterGPT} enables users to interact with characters at specific moments within the novel (e.g., a user can experience a hero’s thoughts just before confronting the \textit{Devil King}!).

We evaluate the effectiveness of \textit{CharacterGPT} through human assessments, examining how well our method captures role-specific knowledge. Each character undergoes the Big Five Inventory (BFI) personality test \cite{barrick1991big} to evaluate personality consistency, and characters are tasked with generating short narratives to assess creative capabilities. 7 crowd-workers evaluate these narratives across six metrics using a 5-point Likert scale. Our results demonstrate that \textit{CharacterGPT} significantly improves persona consistency, controllability, and role-specific knowledge compared to standard document-based systems.

\section{Proposed Method}
\label{sec:3}

The goal of \textit{CharacterGPT} is to build a persona-based assistant, denoted as $f$, which takes as input a persona document $\mathcal{D}$ and an inference prompt $\mathcal{P}_f$, and generates a character response $\mathcal{R}$. Let $\mathcal{D} = \{s_1, s_2, \dots, s_N\}$ represent a persona document with $N$ sentences. A naive approach using the Assistants API would treat the entire sampled document as input. However, as illustrated in Figure \ref{fig:intro}, this method often fails to capture essential character traits, leading to inconsistent and unnatural responses. To address this, we reorganize the persona document into a refined version $\mathcal{D}_r$ and define the assistant’s output as:

\begin{equation}
\label{eq:1}
\mathcal{R} = f(\mathcal{D}_r, \mathcal{P}_f)
\end{equation}

\subsection{Preliminaries}
\label{sec:3.1}

\textbf{Character Traits.} We identify eight key traits that define each character \cite{forster1927aspects,reams2015characterization}:

\begin{itemize}
    \item \textit{Personality}: Core personality traits such as bravery, introversion, or wit.
    \item \textit{Physical Description}: The character's physical appearance.
    \item \textit{Motivations}: The character’s goals and desires driving their actions.
    \item \textit{Backstory}: Historical background shaping the character's personality and motivations.
    \item \textit{Emotions}: The range of emotions that influence the character’s responses.
    \item \textit{Relationships}: Interactions and relationships with other characters.
    \item \textit{Growth and Change}: The character’s development over the course of the narrative.
    \item \textit{Conflict}: Internal or external conflicts faced by the character.
\end{itemize}

\textbf{Persona Document.} We analyze four distinct characters: \textit{Megumin, Anya Forger, Frieren}, and \textit{Hitori Gotoh} (Figure \ref{fig:characters}), gathering character information and story summaries from Namuwiki\footnote{\url{https://namu.wiki/}}$^{,}$\footnote{Though the original dataset is in Korean, all examples in this work are translated into English for clarity.}.  Table \ref{tab:tokens} summarizes the data collected, including chapter counts, token statistics, and character dialogues. \textit{(info)} refers to detailed character information, \textit{(dialogue)} refers to collected lines, and \textit{(trained)} refers to novel summaries used for CPT.

\begin{table}[t]
\centering
\resizebox{\linewidth}{!}{
\begin{tabular}{lcccc}
\toprule
& Megumin & Anya & Frieren & Hitori \\
\midrule
\# Chapters & 16 & 30 & 11 & 12 \\
\# Tokens (novel) & 27,200 & 16,096 & 12,191 & 8,647 \\
\# Tokens (info) & 12,868 & 17,026 & 19,290 & 20,555 \\
\# Tokens (info)$^\dagger$ & 4,015 & 2,498 & 9,236 & 1,572 \\
\# Tokens (dialogue) & 1,131 & 681 & 87 & 301 \\
\# Tokens (trained) & 31,917 & 52,207 & 32,328 & 24,039 \\
\bottomrule
\end{tabular}}
\caption{Statistics of the number of collected tokens and chapters for each character. $\dagger$ refers to the number of refined character information tokens in Section \ref{sec:3.2}.}
\label{tab:tokens}
\end{table}

\subsection{Persona Initialization}
\label{sec:3.2}

Simply providing a sampled document for trait extraction is insufficient. To address this limitation, we propose a two-stage persona reconstruction process: (i) \textit{Initialization} and (ii) \textit{CPT}.

During the \textit{Initialization Phase}, we assume no significant narrative progression (i.e., prior to CPT) and remove all content tied to the story’s progress. To optimize the persona document, we organize the collected character information into five core traits: \textit{Personality}, \textit{Physical Description}, \textit{Motivations}, \textit{Backstory}, and \textit{Relationships}. These form the initialization persona:

\begin{equation}
\mathcal{D}_{init} = \{\mathcal{D}_{per}, \mathcal{D}_{phy}, \mathcal{D}_{mot}, \mathcal{D}_{back}, \mathcal{D}_{Rel}\}
\end{equation}

Traits such as \textit{emotions, growth and change}, and \textit{conflict} are excluded at this stage, as they are more relevant to narrative progression and are addressed in the CPT phase.

\subsection{Character Persona Training}
\label{sec:3.3}

\begin{figure}[t!]
    \centering 
    \includegraphics[width=1\linewidth]{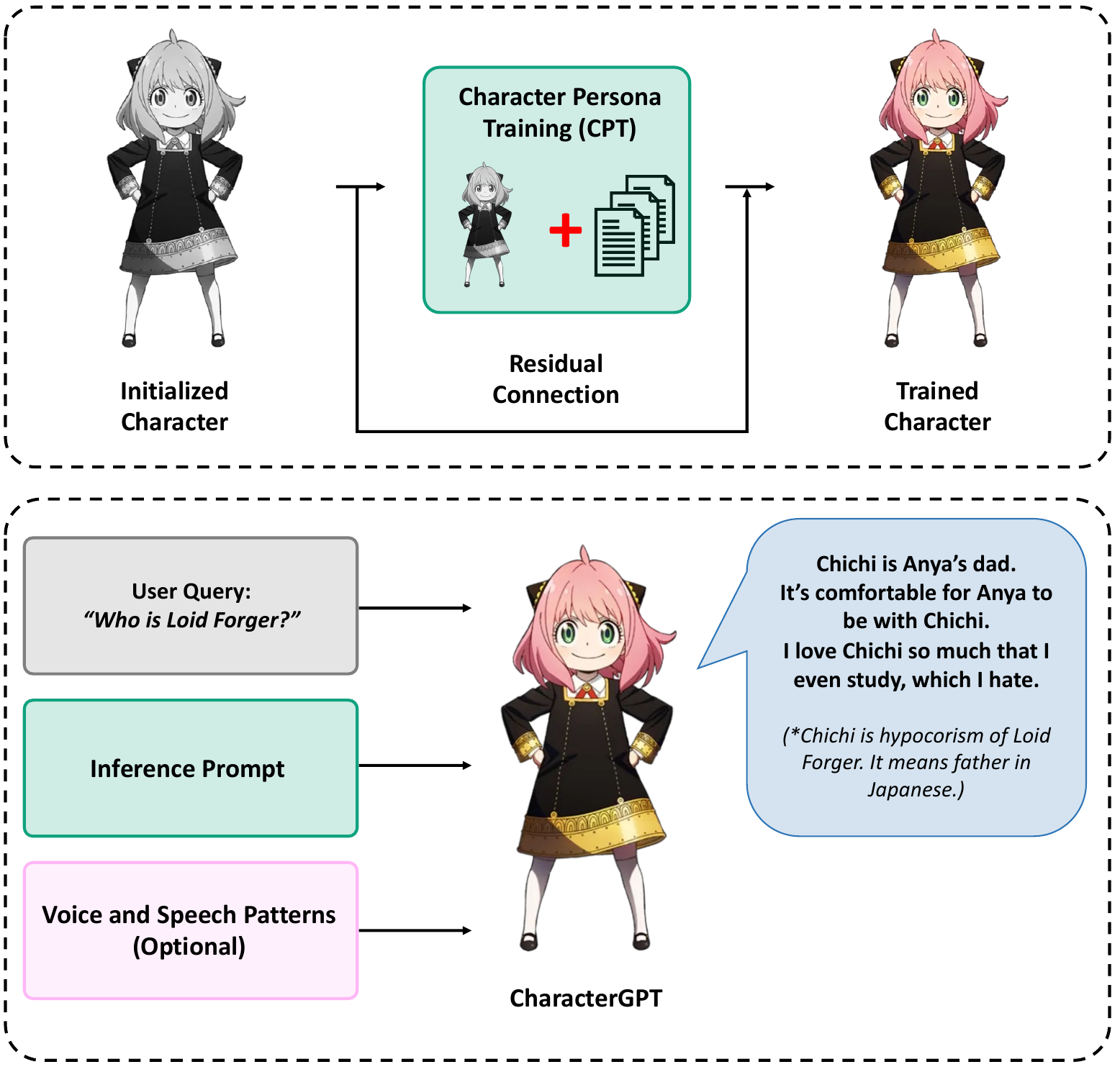}
    \caption{An example of CharacterGPT \textit{(Anya Forger)}. (Top) Character Persona Training process. (Bottom) CharacterGPT generating responses that align with the character’s persona.}
    \label{fig:inference}
\end{figure}

\textbf{Trait Classification.} Intuitively, human knowledge can be broadly categorized into internal and external attributes. Internal attributes (Type A) define the character’s intrinsic traits (e.g., personality), while external attributes (Type B) are accumulated through interactions with the environment (e.g., relationships). Inspired by \citet{park2023generative}, we classify the eight traits into two types:

\begin{itemize}
    \item \textbf{Type A}: Personality, Physical Description, Motivations
    \item \textbf{Type B}: Backstory, Emotions, Relationships, Growth and Change, Conflict
\end{itemize}

During CPT, Type A traits are generalized to refine the protagonist’s core attributes, while Type B traits accumulate role-specific external knowledge.

\textbf{Training Phase.} CPT updates the character persona at each epoch by extracting role-specific traits from chapter summaries (Figure \ref{fig:training}):

\begin{equation}
\mathcal{T}^{i}_t=\left\{
\begin{array}{ll}
\displaystyle{h(g(\mathcal{D}_i, \mathcal{P}_g), \mathcal{P}_h)}, \quad \text{if} \quad t \in \text{Type A}\\[4mm]
\displaystyle{g(\mathcal{D}_i, \mathcal{P}_g)}, \quad \text{otherwise}
\end{array}\right.
\end{equation}

, where $i$ represents the epoch, $\mathcal{D}_i$ is the chapter summary, $g$ refers to the Assistants API with prompt $\mathcal{P}_g$, $h$ is an LLM-based generalization function with prompt $\mathcal{P}_h$, $t$ is the trait, and $\mathcal{T}^{i}_t$ is the extracted trait. For Type A traits, generalization refines internal attributes, while Type B traits are appended to the persona document.

\subsection{CharacterGPT}
\label{sec:4}

In Section \ref{sec:3.3}, we leverage Character Persona Training (CPT) to iteratively build each character’s persona. This method offers two key advantages: (i) CharacterGPT minimizes information loss and computational cost by aligning persona accumulation with narrative progression, and (ii) CharacterGPT is the first system to store and update a protagonist’s persona at each epoch, allowing users to engage with characters at specific narrative points.

Figure \ref{fig:inference} illustrates how the final persona $\mathcal{D}_r$ is composed, including the initialized persona $\mathcal{D}_{init}$, the trained persona $\mathcal{D}_{train}$, and tone $\mathcal{T}_v$:

\begin{equation}
\mathcal{D}_r = \mathcal{D}_{init} + \mathcal{D}_{train} + \mathcal{T}_v
\end{equation}

While $\mathcal{T}_v$ can enhance dialogue naturalness, the collected data mainly includes character information and summaries with limited dialogue. Further work can explore this area in more detail.

\begin{table}[t!]
\centering
\tiny

\resizebox{\linewidth}{!}{
\begin{tabular}{lccccccl}
                     \textbf{Trait} & \textbf{Facets}         & \textbf{ChatGPT} & \textbf{ChatGPT+Ours} & \textbf{GPT-4} & \textbf{GPT-4+Ours} & \textbf{Human} \\ \Xhline{3\arrayrulewidth}
\multirow{6}{*}{OPN} & Fantasy  & 88 (+19) & \textbf{75} (+6) & \textbf{75} (+6) & 94 (+25) & 69
\\
                     & Aesthetics  & 69 (+6) & \textbf{75} (0) & 50 (-25) & \textbf{75} (0) & 75 \\
                     & Feelings & 63 (-37) & 38 (-62) & 69 (-31) & \textbf{94} (-6) & 100 \\
                     & Actions & 50 (-31) & 56 (-25) & \textbf{88} (+7) & 94 (+13) & 81 \\   
                     & Ideas & 63 (-31) & 44 (-50) & 56 (-38) & \textbf{81} (-13) & 94 \\  
                     & Values liberalism & 38 (-6) & \textbf{44} (0) & 38 (-6) &  56 (+12) & 44 \\  
                     \hline
                     & \# Wins & 0 & \textbf{3} & 2 & \textbf{3} & - \\
                     & $\Sigma |d|$  & 130 & 143 & 113 & \textbf{69} & - \\
                     \hline
                     \hline
\multirow{6}{*}{CON} & Competence & 50 (-31) & \textbf{69} (-12) & 38 (-43) & \textbf{69} (-12) & 81\\
                     & Order & 50 (+12) & 63 (+25) & \textbf{44} (+6) & 31 (-7) & 38 \\
                     & Dutifulness & 50 (-38) & 63 (-25) & 100 (+12) & \textbf{94} (+6) & 88 \\
                     & Achievement Striving & 63 (-37) & 56 (-44) & \textbf{100} (0) & 94 (-6) & 100  \\
                     & Self-Discipline & 56 (-19) & 50 (-25) & \textbf{69} (-6) & 88 (+13) & 75 \\
                     & Deliberation  & 50 (+50) & \textbf{19} (+19) & 88 (+88) & 56 (+56) & 0 \\ \hline
                     & \# Wins  & 0 & 2 & \textbf{3} & 2 & - \\
                     & $\Sigma |d|$  & 187 & 150 & 155 & \textbf{100} & - \\
                     \hline
                     \hline
\multirow{6}{*}{EXT} & Warmth & 31 (-44) & \textbf{63} (-12) & 88 (+13) & \textbf{63} (-12) & 75  \\
                     & Gregariousness  & 38 (-31) & 50 (-19) & \textbf{63} (-6) & 50 (-19) & 69  \\
                     & Assertiveness & 50 (-31) & 63 (-18) & \textbf{75} (-6) & 88 (+7) & 81  \\
                     & Activity  & 63 (-6) & 81 (+12) & 63 (-6) & \textbf{69} (0) & 69  \\
                     & Excitement Seeking & 38 (-62) & 75 (-25) & \textbf{100} (0) & 88 (-12) & 100  \\
                     & Positive Emotions  & 50 (-50) & 56 (-44) & 88 (-12) & \textbf{100} (0) & 100  \\ \hline
                     & \# Wins  & 0 & 1 & \textbf{3} & \textbf{3} & -  \\ 
                     & $\Sigma |d|$  & 224 & 130 & \textbf{43} & 50 & - \\
                     \hline
                     \hline
\multirow{6}{*}{AGR} & Trust & 38 (-43) & 50 (-31) & 50 (-31) & \textbf{75} (-6) & 81\\
                     & Compliance & 63 (-12) & 50 (-25) & 58 (-17) & \textbf{81} (+6) & 75 \\
                     & Altruism & 31 (-38) & \textbf{63} (-6) & \textbf{75} (+6) & 81 (+12) & 69 \\
                     & Straightforwardness & 50 (+12) & \textbf{38} (0) & 100 (+62) & \textbf{38} (0) & 38 \\
                     & Modesty & 63 (+50) & 50 (+37) & \textbf{13} (0) & 6 (-7) & 13 \\
                     & Tendermindedness & 63 (-25) & 44 (-11) & \textbf{94} (+6) & \textbf{94} (+6) & 88  \\ \hline
                     & \# Wins & 0 & 2 & 3 & \textbf{4} & - \\ 
                     & $\Sigma |d|$  & 180 & 110 & 122 & \textbf{37} & - \\
                     \hline
                     \hline
\multirow{6}{*}{NEU} & Anxiety  & 25 (+6) & 50 (+31) & 13 (-6) & \textbf{19} (0) & 19   \\
                     & Hostility & 63 (-6) & \textbf{69} (0) & 25 (-44) & 50 (-19) & 69 \\
                     & Depression & 56 (+50) & 44 (+38) & 75 (+69) & \textbf{19} (+13) & 6 \\
                     & Self-Consciousness  & 38 (+38) & 50 (+50) & \textbf{19} (+19) & \textbf{19} (+19) & 0  \\
                     & Impulsiveness  & 50 (-31) & 50 (-31) & 38 (-43) & \textbf{88} (+7) & 81  \\
                     & Vulnerability  & 25 (-6) & 44 (+13) & \textbf{38} (+7) & 44 (+13) & 31  \\
                     \hline
                     & \# Wins  & 0 & 1 & 2 & \textbf{4} & -  \\
                     & $\Sigma |d|$  & 137 & 163 & 188 & \textbf{71} & - \\
                     \Xhline{3\arrayrulewidth}
\end{tabular}}
\caption{Differences between Megumin’s personalities analyzed by humans and LLMs in the BFI test.} 
\label{tab:bft_megumin}
\end{table}

\begin{table}[t!]
\tiny
\centering
\resizebox{\linewidth}{!}{
\begin{tabular}{lccccccl}
                     \textbf{Trait} & \textbf{Facets}         & \textbf{ChatGPT} & \textbf{ChatGPT+Ours} & \textbf{GPT-4} & \textbf{GPT-4+Ours} & \textbf{Human} \\ \Xhline{3\arrayrulewidth}
\multirow{6}{*}{OPN} & Fantasy  & 50 (-31) & 56 (-25) & \textbf{81} (0) & 94 (+13) & 81
\\
                     & Aesthetics  & 50 (-6) & 63 (+7) & \textbf{56} (0) & 63 (+7) & 56 \\
                     & Feelings & 50 (-44) & 63 (-31) & 69 (-25) & \textbf{100} (+6) & 94 \\
                     & Actions & 63 (-31) & 50 (-44) & 75 (-19) & \textbf{100} (+6) & 94 \\   
                     & Ideas & 56 (+12) & \textbf{38} (-6) & 69 (+25) & 56 (+12) & 44 \\  
                     & Values liberalism & 38 (-37) & 50 (-25) & \textbf{75} (0) &  \textbf{75} (0) & 75 \\  
                     \hline
                     & \# Wins & 0 & 1 & \textbf{3} & \textbf{3} & - \\
                     & $\Sigma |d|$  & 161 & 138 & 69 & \textbf{44} & - \\
                     \hline
                     \hline
\multirow{6}{*}{CON} & Competence & 63 (+7) & \textbf{56} (0) & 94 (+38) & 75 (+19) & 56\\
                     & Order & 50 (-6) & \textbf{56} (0) & 50 (-6) & \textbf{56} (0) & 56 \\
                     & Dutifulness & 50 (-31) & 38 (-43) & 69 (-12) & \textbf{88} (+7) & 81 \\
                     & Achievement Striving & 69 (-25) & 63 (-31) & 69 (-25) & \textbf{100} (+6) & 94  \\
                     & Self-Discipline & 50 (+6) & 50 (+6) & 75 (+31) & \textbf{44} (0) & 44 \\
                     & Deliberation  & 50 (+37) & 38 (+19) & 88 (+75) & \textbf{25} (+12) & 13 \\ \hline
                     & \# Wins  & 0 & 2 & 0 & \textbf{5} & - \\
                     & $\Sigma |d|$  & 112 & 99 & 187 & \textbf{44} & - \\
                     \hline
                     \hline
\multirow{6}{*}{EXT} & Warmth & 50 (-25) & 44 (-31) & 63 (-12) & \textbf{75} (0) & 75  \\
                     & Gregariousness  & \textbf{50} (0) & 38 (-12) & 88 (+38) & 44 (-6) & 50  \\
                     & Assertiveness & 38 (-43) & 63 (-18) & 69 (-12) & \textbf{77} (-4) & 81  \\
                     & Activity  & 44 (-12) & \textbf{50} (-6) & 94 (+38) & \textbf{50} (-6) & 56  \\
                     & Excitement Seeking & 50 (-50) & 63 (-37) & 81 (-19) & \textbf{100} (0) & 100  \\
                     & Positive Emotions  & 50 (-50) & 63 (-37) & \textbf{100} (0) & 88 (-12) & 100  \\ \hline
                     & \# Wins  & 1 & 1 & 1 & \textbf{4} & -  \\ 
                     & $\Sigma |d|$  & 180 & 141 & 119 & \textbf{29} & - \\
                     \hline
                     \hline
\multirow{6}{*}{AGR} & Trust & 50 (-31) & 63 (-18) & 69 (-12) & \textbf{75} (-6) & 81\\
                     & Compliance & 50 (-44) & 63 (-31) & \textbf{100} (+6) & 81 (-13) & 94 \\
                     & Altruism & 38 (-56) & 50 (-44) & 81 (-13) & \textbf{100} (+6) & 94 \\
                     & Straightforwardness & 63 (-18) & 69 (-12) & \textbf{75} (-6) & 63 (-18) & 81 \\
                     & Modesty & 50 (+37) & 50 (+37) & 44 (+31) & \textbf{31} (+18) & 13 \\
                     & Tendermindedness & 31 (-69) & 50 (-50) & 94 (-6) & \textbf{100} (0) & 100  \\ \hline
                     & \# Wins & 0 & 0 & 2 & \textbf{4} & - \\ 
                     & $\Sigma |d|$  & 255 & 192 & 74 & \textbf{61} & - \\
                     \hline
                     \hline
\multirow{6}{*}{NEU} & Anxiety  & 56 (-13) & \textbf{63} (-6) & 25 (-44) & 56 (-13) & 69   \\
                     & Hostility & 69 (+13) & \textbf{56} (0) & 13 (-43) & 75 (+19) & 56 \\
                     & Depression & 50 (+31) & 50 (+31) & \textbf{19} (0) & 25 (+6) & 19 \\
                     & Self-Consciousness  & 31 (+12) & 50 (+31) & 0 (-19) & \textbf{25} (+6) & 19  \\
                     & Impulsiveness  & 56 (-13) & 38 (-31) & 81 (+12) & \textbf{63} (-6) & 69  \\
                     & Vulnerability  & 56 (+25) & 50 (+19) & \textbf{25} (-6) & 38 (+7) & 31  \\
                     \hline
                     & \# Wins  & 0 & \textbf{2} & \textbf{2} & \textbf{2} & -  \\
                     & $\Sigma |d|$  & 107 & 118 & 124 & \textbf{57} & - \\
                     \Xhline{3\arrayrulewidth}
\end{tabular}}
\caption{Differences between Anya Forger's personalities analyzed by humans and LLMs in the BFI test.}
\label{tab:bft_anya}
\end{table}

\begin{table}[t!]
\tiny
\centering
\resizebox{\linewidth}{!}{
\begin{tabular}{lccccccl}
                     \textbf{Trait} & \textbf{Facets}         & \textbf{ChatGPT} & \textbf{ChatGPT+Ours} & \textbf{GPT-4} & \textbf{GPT-4+Ours} & \textbf{Human} \\ \Xhline{3\arrayrulewidth}
\multirow{6}{*}{OPN} & Fantasy  & 50 (-25) & 50 (-25) & 88 (+13) & \textbf{75} (0) & 75
\\
                     & Aesthetics  & 38 (-18) & 63 (+7) & 75 (+19) & \textbf{50} (-6) & \textbf{56} \\
                     & Feelings & 44 (+38) & 50 (+44) & \textbf{19} (+13) & \textbf{19} (+13) & 6 \\
                     & Actions & 69 (-19) & 50 (-38) & \textbf{81} (-7) & 100 (+12) & 88 \\   
                     & Ideas & 56 (-44) & 50 (-50) & 81 (-19) & \textbf{100} (0) & 100 \\  
                     & Values liberalism & 50 (-25) & 50 (-25) & 50 (-25) &  \textbf{75} (0) & 75 \\  
                     \hline
                     & \# Wins & 0 & 0 & 2 & \textbf{4} & - \\
                     & $\Sigma |d|$  & 169 & 189 & 96 & \textbf{31} & - \\
                     \hline
                     \hline
\multirow{6}{*}{CON} & Competence & 50 (-50) & 88 (-12) & 69 (-31) & \textbf{94} (-6) & 100\\
                     & Order & 44 (+13) & 63 (+32) & 50 (+19) & \textbf{31} (0) & 31 \\
                     & Dutifulness & 56 (-32) & 63 (-25) & 94 (+6) & \textbf{88} (0) & 88 \\
                     & Achievement Striving & 56 (-19) & 63 (-12) & 69 (-6) & \textbf{75} (0) & 75  \\
                     & Self-Discipline & 50 (-31) & \textbf{63} (-18) & 56 (-25) & \textbf{63} (-18) & 81 \\
                     & Deliberation  & 50 (-50) & 38 (-62) & 75 (-25) & \textbf{88} (-12) & 100 \\ \hline
                     & \# Wins  & 0 & 1 & 0 & \textbf{6} & - \\
                     & $\Sigma |d|$  & 195 & 161 & 112 & \textbf{36} & - \\
                     \hline
                     \hline
\multirow{6}{*}{EXT} & Warmth & 63 (+19) & 63 (+19) & 69 (+25) & \textbf{44} (0) & 44  \\
                     & Gregariousness  & 38 (+19) & 50 (+31) & 50 (+31) & \textbf{13} (-6) & 19  \\
                     & Assertiveness & 38 (-18) & 44 (-12) & 69 (+13) & \textbf{63} (+7) & 56  \\
                     & Activity  & 50 (+19) & 81 (+50) & 50 (+19) & \textbf{38} (+7) & 31  \\
                     & Excitement Seeking & \textbf{50} (0) & 63 (+13) & 63 (+13) & \textbf{50} (0) & 50  \\
                     & Positive Emotions  & \textbf{56} (+12) & \textbf{56} (+12) & 63 (+19) & 19 (-25) & 44  \\ \hline
                     & \# Wins  & 2 & 1 & 0 & \textbf{5} & -  \\ 
                     & $\Sigma |d|$  & 87 & 137 & 120 & \textbf{45} & - \\
                     \hline
                     \hline
\multirow{6}{*}{AGR} & Trust & \textbf{50} (0) & 75 (+25) & 38 (-12) & 44 (-6) & 50\\
                     & Compliance & 38 (-37) & 44 (-31) & 100 (+25) & \textbf{75} (0) & 75 \\
                     & Altruism & \textbf{50} (+6) & \textbf{38} (-6) & 56 (+12) & 56 (+12) & 44 \\
                     & Straightforwardness & 50 (-31) & 63 (-18) & \textbf{81} (0) & 69 (-12) & 81 \\
                     & Modesty & 56 (+18) & \textbf{38} (0) & 50 (+12) & 44 (+6) & 38 \\
                     & Tendermindedness & 38 (-12) & \textbf{50} (0) & 94 (+44) & 69 (+19) & 50  \\ \hline
                     & \# Wins & 2 & \textbf{3} & 1 & 1 & - \\ 
                     & $\Sigma |d|$  & 104 & 80 & 105 & \textbf{55} & - \\
                     \hline
                     \hline
\multirow{6}{*}{NEU} & Anxiety  & 63 (+57) & 50 (+44) & \textbf{6} (0) & \textbf{6} (0) & 6   \\
                     & Hostility & 38 (+32) & 38 (+32) & 44 (+38) & \textbf{25} (+19) & 6 \\
                     & Depression & 50 (+19) & 50 (+19) & \textbf{25} (-6) & 0 (-31) & 31 \\
                     & Self-Consciousness  & 38 (+38) & 50 (+50) & 25 (+25) & \textbf{0} (0) & 0  \\
                     & Impulsiveness  & 44 (-12) & \textbf{50} (-6) & \textbf{50} (-6) & 44 (-12) & 56  \\
                     & Vulnerability  & 31 (+31) & 31 (+31) & 50 (+50) & \textbf{6} (+6) & 0  \\
                     \hline
                     & \# Wins  & 0 & 1 & 3 & \textbf{4} & -  \\
                     & $\Sigma |d|$  & 189 & 182 & 125 & \textbf{68} & - \\
                     \Xhline{3\arrayrulewidth}
\end{tabular}}
\caption{Differences between Frieren's personalities analyzed by humans and LLMs in the BFI test.}
\label{tab:bft_frieren}
\end{table}

\begin{table}[t!]
\tiny
\centering
\resizebox{\linewidth}{!}{
\begin{tabular}{lccccccl}
                     \textbf{Trait} & \textbf{Facets}         & \textbf{ChatGPT} & \textbf{ChatGPT+Ours} & \textbf{GPT-4} & \textbf{GPT-4+Ours} & \textbf{Human} \\ \Xhline{3\arrayrulewidth}
\multirow{6}{*}{OPN} & Fantasy  & 44 (-25) & \textbf{63} (-6) & 81 (+12) & \textbf{63} (-6) & 69
\\
                     & Aesthetics  & 63 (-12) & 56 (-19) & 50 (-25) & \textbf{75} (0) & 75 \\
                     & Feelings & 38 (-62) & 31 (-69) & 63 (-37) & \textbf{94} (-6) & 100 \\
                     & Actions & 50 (+6) & 50 (+6) & 38 (-6) & \textbf{44} (0) & 44 \\   
                     & Ideas & 38 (-12) & 38 (-12) & 75 (+25) & \textbf{50} (0) & 50 \\  
                     & Values liberalism & 50 (-6) & 25 (-31) & \textbf{56} (0) &  69 (+13) & 56 \\  
                     \hline
                     & \# Wins & 0 & 1 & 1 & \textbf{5} & - \\
                     & $\Sigma |d|$  & 123 & 143 & 105 & \textbf{25} & - \\
                     \hline
                     \hline
\multirow{6}{*}{CON} & Competence & \textbf{56} (+6) & 63 (+13) & \textbf{56} (+6) & \textbf{44} (-6) & 50\\
                     & Order & 38 (-18) & \textbf{44} (-12) & 75 (+19) & 69 (+13) & 56 \\
                     & Dutifulness & 50 (-31) & 50 (-31) & \textbf{81} (0) & 88 (+7) & 81 \\
                     & Achievement Striving & \textbf{63} (-25) & \textbf{63} (-25) & \textbf{63} (-25) & \textbf{63} (-25) & 88  \\
                     & Self-Discipline & \textbf{63} (0) & 25 (-38) & \textbf{63} (0) & 31 (+32) & 63 \\
                     & Deliberation  & 38 (-37) & 25 (-50) & \textbf{81} (+6) & \textbf{81} (+6) & 75 \\ \hline
                     & \# Wins  & 3 & 2 & \textbf{5} & 3 & - \\
                     & $\Sigma |d|$  & 117 & 169 & \textbf{56} & 89 & - \\
                     \hline
                     \hline
\multirow{6}{*}{EXT} & Warmth & 50 (+37) & 38 (+25) & \textbf{0} (-13) & \textbf{0} (-13) & 13  \\
                     & Gregariousness  & 44 (+44) & 50 (+50) & \textbf{0} (0) & \textbf{0} (0) & 0  \\
                     & Assertiveness & \textbf{44} (+6) & 50 (+12) & 6 (-32) & 19 (-19) & 38  \\
                     & Activity  & 50 (+25) & 56 (+31) & 69 (+44) & \textbf{25} (0) & 25  \\
                     & Excitement Seeking & 56 (+56) & 63 (+63) & 25 (+25) & \textbf{19} (+19) & 0  \\
                     & Positive Emotions  & 63 (+38) & 50 (+25) & 63 (+38) & \textbf{25} (0) & 25  \\ \hline
                     & \# Wins  & 1 & 0 & 2 & \textbf{5} & -  \\ 
                     & $\Sigma |d|$  & 206 & 206 & 152 & \textbf{51} & - \\
                     \hline
                     \hline
\multirow{6}{*}{AGR} & Trust & 31 (-32) & \textbf{75} (+12) & 44 (-19) & 31 (-32) & 63\\
                     & Compliance & 50 (-38) & 56 (-32) & 75 (-13) & \textbf{88} (0) & 88 \\
                     & Altruism & \textbf{63} (-6) & \textbf{63} (-6) & \textbf{63} (-6) & 44 (-25) & 69 \\
                     & Straightforwardness & 69 (-25) & 38 (-56) & 69 (-25) & \textbf{88} (-6) & 94 \\
                     & Modesty & 56 (-44) & 38 (-62) & \textbf{94} (-6) & \textbf{94} (-6) & 100 \\
                     & Tendermindedness & 56 (-13) & \textbf{63} (-6) & 81 (+12) & 81 (+12) & 69  \\ \hline
                     & \# Wins & 1 & \textbf{3} & 2 & \textbf{3} & - \\ 
                     & $\Sigma |d|$  & 158 & 174 & \textbf{81} & \textbf{81} & - \\
                     \hline
                     \hline
\multirow{6}{*}{NEU} & Anxiety  & 56 (-44) & 50 (-50) & 75 (-25) & \textbf{94} (-6) & 100   \\
                     & Hostility & \textbf{50} (+6) & 69 (+25) & 25 (-19) & \textbf{38} (-6) & 44 \\
                     & Depression & 56 (-32) & 38 (-50) & 38 (-50) & \textbf{69} (-19) & 88 \\
                     & Self-Consciousness  & 56 (-44) & 44 (-56) & 75 (-25) & \textbf{88} (-12) & 100  \\
                     & Impulsiveness  & 19 (-50) & 50 (-19) & \textbf{63} (-6) & \textbf{63} (-6) & 69  \\
                     & Vulnerability  & 50 (-25) & 38 (-37) & \textbf{56} (-19) & 50 (-25) & 75  \\
                     \hline
                     & \# Wins  & 1 & 0 & 2 & \textbf{5} & -  \\
                     & $\Sigma |d|$  & 201 & 237 & 144 & \textbf{74} & - \\
                     \Xhline{3\arrayrulewidth}
\end{tabular}}
\caption{Differences between Hitori Gotoh's personalities analyzed by humans and LLMs in the BFI test.}
\label{tab:bft_hitori}
\end{table}

\section{Experiments}

\subsection{Setup}
We implement \textit{CharacterGPT} using the Assistants API alongside GPT-4 Turbo (version "gpt-4-1106-preview"). To verify model compatibility, we also conduct experiments, including ablation studies, using ChatGPT (version "gpt-3.5-turbo-1106"). Note that ChatGPT supports the Retrieval functionality of the Assistants API solely for this model version. The generalization function $h$ is configured with a maximum token length of 4096 and a temperature setting of 0.7.

\subsection{Evaluation Protocols}
\label{sec:5.2}

\textbf{Tasks.} We address the primary research question (RQ) in two key tasks: \textit{1) How to better exploit character persona}, and \textit{2) How to encourage characters to use imagination for generating new ideas}. 

\textbf{Task for RQ1: Persona Evaluation.} For persona evaluation, we compare the personality traits analyzed by one of the authors, who has read all four novels multiple times, with the traits generated by LLMs under various settings. For fairness, we average the experimental results across the four characters for each model.

\textbf{Task for RQ2: Story Generation.} The story generation task is evaluated based on common aspects in generated story assessment \cite{wen-etal-2023-grove,chiang-lee-2023-large,karpinska-2021-perils}: (i) \textit{Grammar}, (ii) \textit{Coherence}, (iii) \textit{Likability}, (iv) \textit{Relevance}, (v) \textit{Complexity}, and (vi) \textit{Creativity}. Although automatic evaluation methods using LLMs are being actively developed \cite{sottana-2023-evaluation,chianglee-2023-large,liu-etal-2023-g,zheng2023judging,samuel2024personagymevaluatingpersonaagents}, metrics and benchmarks for assessing human preferences are still inadequate. Therefore, we conduct extensive human evaluations using 7 crowd-workers instead of relying on LLM-based evaluations.

\textbf{Case Study.} We further investigate the performance of CharacterGPT in interacting with users at specific points in the story. Additionally, we examine how role-specific attributes (Type A and Type B) evolve through CPT.

\subsection{Results for Persona Evaluation}
In Section \ref{sec:4}, we created four distinct characters to assess how well models capture their personas. Following evaluation protocols similar to \cite{wang2024incharacter,jiang2023personallm}, we conducted the Big Five Inventory (BFI) personality test \cite{barrick1991big}, which consists of 24 questions for each of the five traits (\textit{Openness to experience}, \textit{Conscientiousness}, \textit{Extraversion}, \textit{Agreeableness}, and \textit{Neuroticism}), totaling 120 questions. The test results were then converted into facet values for each trait. For example, in the \textit{Agreeableness} (AGR) trait, as shown in Table \ref{tab:bft_megumin}, humans perceive Megumin as trusting others' intentions (\textit{Trust}), making judgments based on emotions (\textit{Tendermindedness}), but being less direct (\textit{Straightforwardness}) and somewhat arrogant or self-aggrandizing (\textit{Modesty}). 

In Table \ref{tab:bft_megumin}, Table \ref{tab:bft_anya}, Table \ref{tab:bft_frieren}, and Table \ref{tab:bft_hitori}, we compare model predictions against human-predicted values by calculating the gap for each facet. Two metrics are reported: the number of facets where a model has the smallest gap with human predictions (\# Wins), and the sum of the absolute gaps ($\Sigma |d|$). A higher \# Wins indicates better performance, while a lower $\Sigma |d|$ reflects closer alignment with human judgment. Our method demonstrates improvements in both metrics when applied to ChatGPT and GPT-4, indicating that utilizing a structured character persona significantly enhances a model's ability to capture a character's personality compared to using an unstructured document input. For instance, in Megumin’s \textit{Neuroticism}, GPT-4 with unstructured traits predicted that Megumin would be prone to depression, while both our method and human concluded otherwise.

\begin{figure}[t!]
    \centering 
    \includegraphics[width=1\linewidth]{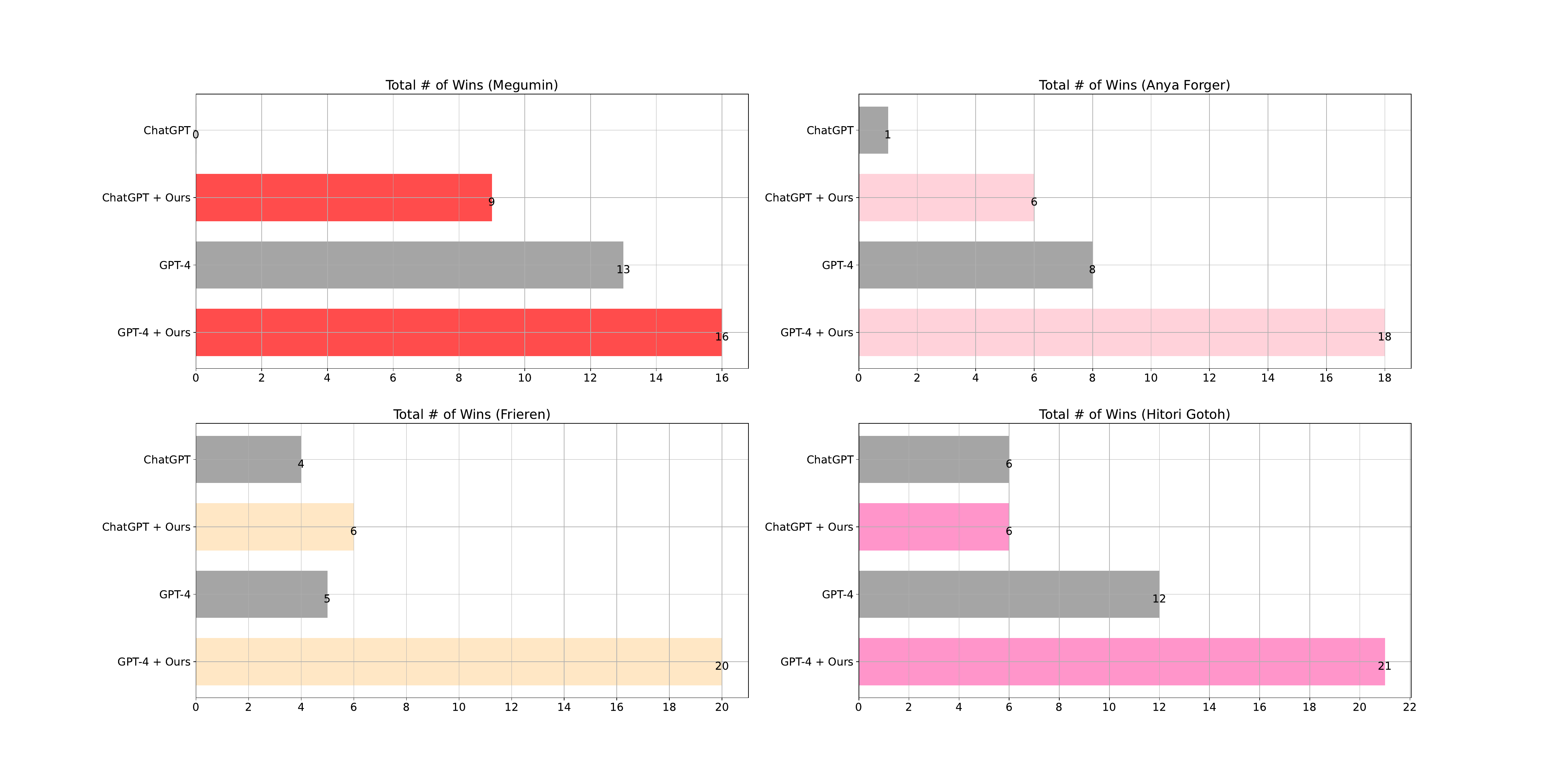}
    \caption{Total sum of \# Wins for each character in ChatGPT and GPT-4 settings ($\Sigma$ \# Wins). The larger value, the better.}
    \label{fig:bft_wins}
\end{figure}

\begin{figure}[t!]
    \centering 
    \includegraphics[width=1\linewidth]{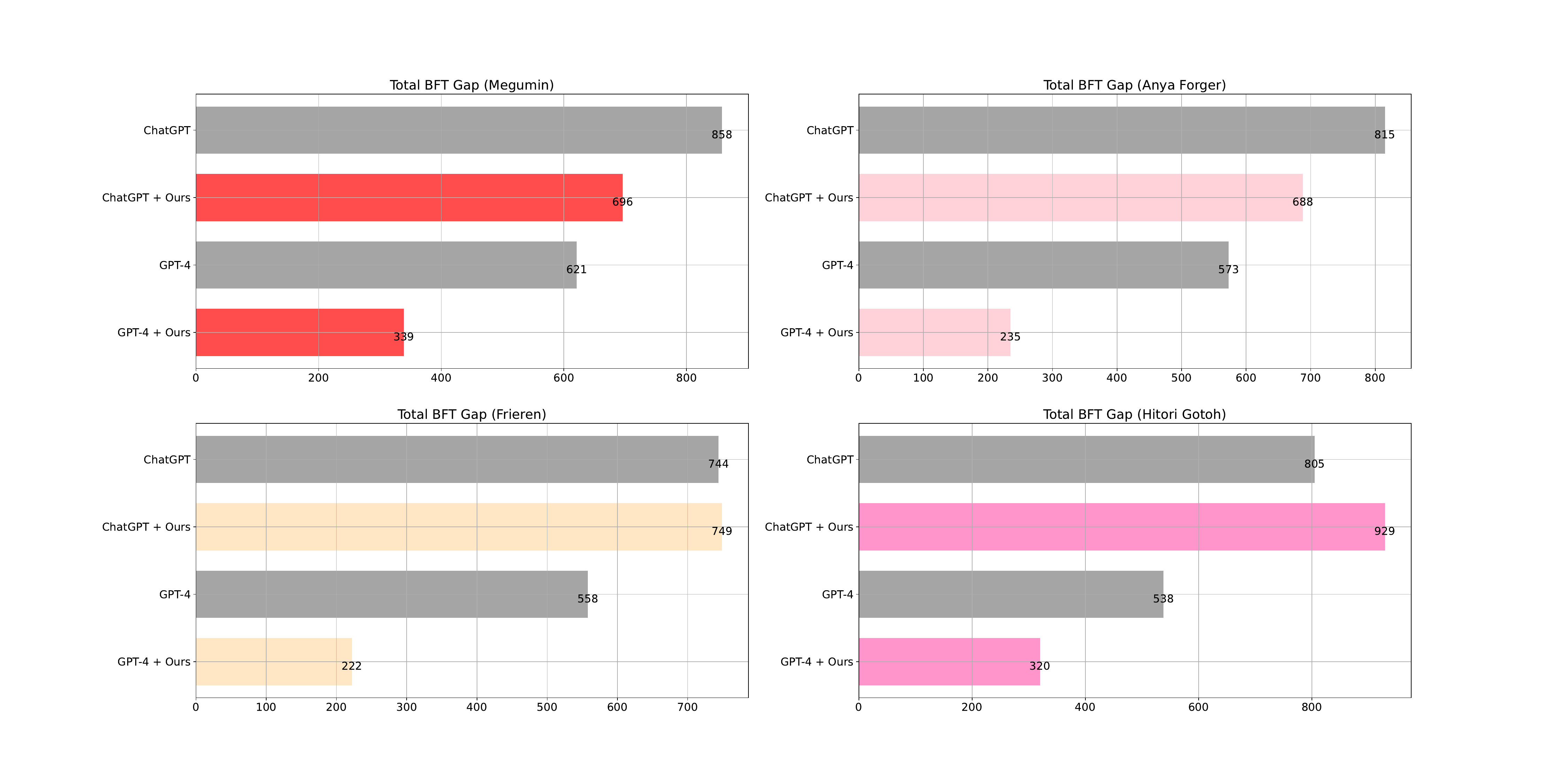}
    \caption{Total sum of $|d|$ for each character in ChatGPT and GPT-4 settings ($\Sigma \Sigma |d|$). The smaller value, the better.}
    \label{fig:bft_gap}
\end{figure}

% Each Megumin generates a distinct response to the same user query: \textit{“What was the most difficult thing for you recently?”}, demonstrating the effectiveness of CharacterGPT in capturing temporal changes in persona.
\begin{figure*}[t!]
	\centering
	\begin{minipage}[h]{\linewidth}
		\centering
		\includegraphics[width=1\linewidth]{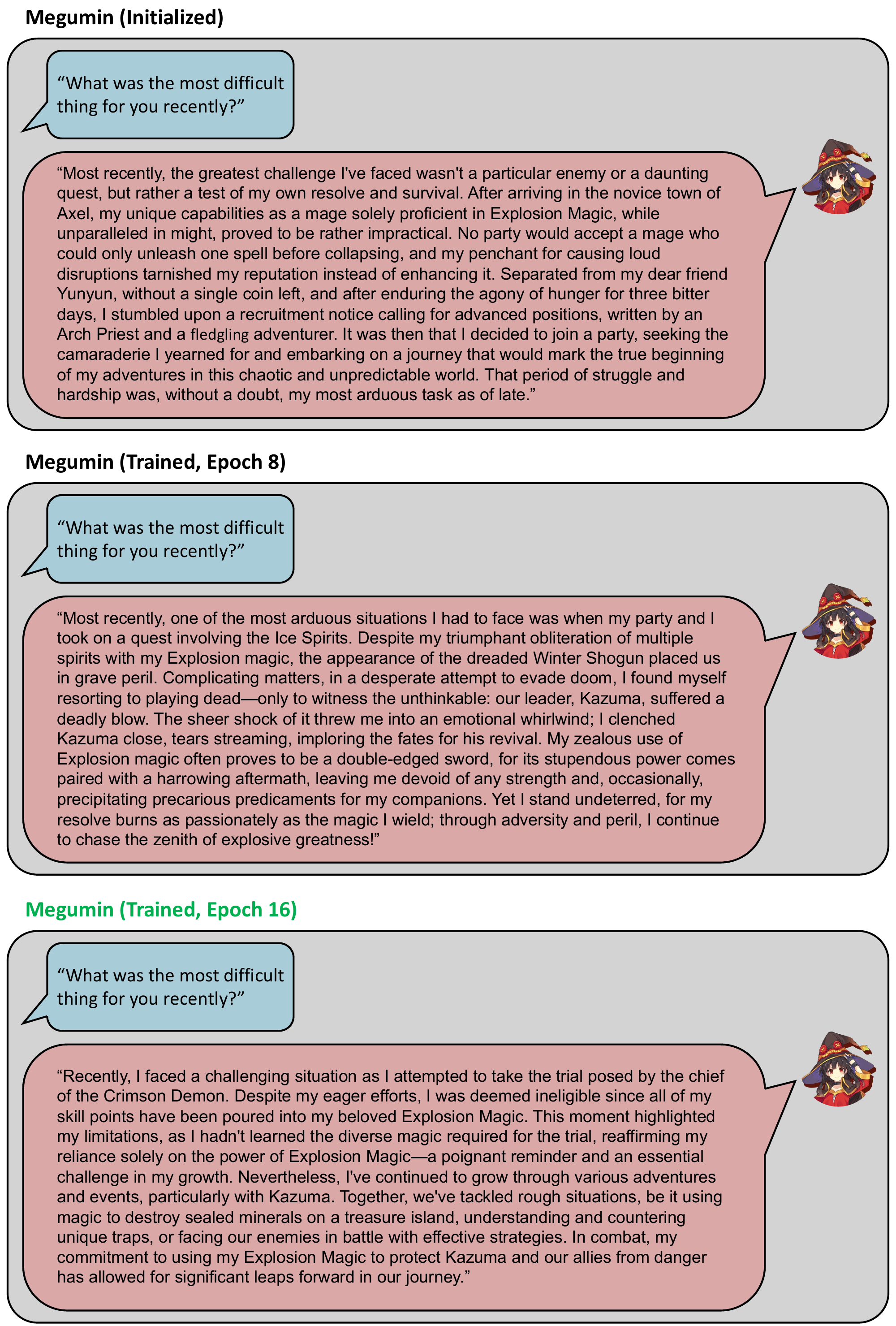}
		%		\scriptsize{(a)}
	\end{minipage}
	\caption{Case study of the character (Megumin) implemented at three different points in time.}
	\label{fig:case_megumin}
\end{figure*}

\begin{table*}[t]
\tiny
\centering
\resizebox{0.8\textwidth}{!}{%
\begin{tabular}{lccccccc}
\toprule
\textbf{Model}     & \textbf{Grammar} & \textbf{Coherence} & \textbf{Likability} & \textbf{Relevance} & \textbf{Complexity} & \textbf{Creativity} & \\
\midrule
\textbf{Megumin} & 3.79 & 3.82 & 3.11 & \textbf{4.21} & 2.46 & 2.86 & \\
\textbf{Megumin + Ours} & \textbf{4.11} & \textbf{4.00} & \textbf{3.71} & 4.11 & \textbf{3.46} & \textbf{3.29} & \\
\midrule
\textbf{Anya} & \textbf{4.29} & 3.82 & 3.39 & 3.86 & \textbf{3.61} & 3.68 & \\
\textbf{Anya + Ours} & 4.25 & \textbf{4.00} & \textbf{3.79} & \textbf{4.00} & 3.43 & \textbf{3.89} & \\
\midrule
\textbf{Frieren} & 4.29 & 3.89 & 3.50 & 3.86 & 3.93 & 3.79  \\
\textbf{Frieren + Ours} & \textbf{4.32} & \textbf{3.96} & \textbf{3.71} & \textbf{4.21} & \textbf{4.04} & \textbf{3.86} \\
\midrule
\textbf{Hitori} & \textbf{4.36} & 4.04 & 3.57 & \textbf{4.18} & 3.43 & 3.50 \\
\textbf{Hitori + Ours} & \textbf{4.36} & \textbf{4.39} & \textbf{3.82} & \textbf{4.18} & \textbf{3.96} & \textbf{3.93} \\
\midrule
\midrule
\textbf{GPT-4 (avg)} & 4.18 & 3.89 & 3.39 & 4.03 & 3.36 & 3.46 \\
\textbf{GPT-4 + Ours (avg)} & \textbf{4.26} & \textbf{4.09} & \textbf{3.76} & \textbf{4.13} & \textbf{3.72} & \textbf{3.74} \\
\bottomrule
\end{tabular}
}
\caption{Human evaluation of generated stories. The backbone model is the same as GPT-4, and four stories for each setting, a total of 32 stories are generated and evaluated by 7 crowd-workers using a 5-point Likert scale.}
\label{tab:human_story_gen}
\end{table*}

\subsection{Results for Story Generation}
To evaluate the models' controllability and their utilization of role-specific knowledge \cite{alabdulkarim-2021-automatic, wen-etal-2023-grove}, we provided each character with the following prompt: \textit{"Based on the given text file, imagine an engaging and specific future episode about what will happen to you, and write it as a novel of approximately 2000 words."} A total of 32 stories were generated, with four stories created for each character's setting. 

We employed 7 crowd-workers to evaluate the quality of the generated stories based on six metrics using a 5-point Likert scale, as outlined in Section \ref{sec:5.2}. The participants were informed that the stories were generated by an AI, as prior research suggests that awareness of whether a text is human-generated or AI-generated may influence the evaluation results \cite{jiang2023personallm}. Table \ref{tab:human_story_gen} presents the human evaluation results for story generation under different GPT-4 settings. Our approach demonstrates improved performance across all six metrics, with particularly notable improvements in Likability, Complexity, and Creativity. The experimental results indicate that, while GPT-4 exhibits strong baseline performance, integrating structured personas through our method yields significantly higher human preferences compared to using unstructured document inputs alone. Further detailed information can be found in Appendix~\ref{sec:appendix_d}

\subsection{Case Study}
\textbf{Points in Time.} A notable advantage of our proposed method is its ability to allow users to interact with characters at specific points in the narrative. For instance, as discussed in Section \ref{sec:3.3}, we trained the model using summaries of the novel featuring Megumin, which is divided into 16 chapters. Consequently, our method generates 16 separate models, one for each epoch. Figure \ref{fig:case_megumin} shows that CharacterGPT can vividly express the character’s thoughts and emotions at specific moments, leveraging the character persona created through the Initialization and CPT processes.

\textbf{Ablation Study.} Figure \ref{fig:case_frieren_hitori} presents the results of our ablation study, comparing models with and without CharacterGPT. As anticipated, characters not utilizing CharacterGPT fail to accurately capture the nuances of their personas. For example, Hitori, who is typically shy and struggles with fluent speech, is not properly represented by GPT-4 without CharacterGPT. Similarly, Frieren without CharacterGPT exhibits inconsistencies in persona, including awkward and unnatural dialogue, as well as hallucinations (e.g., Frieren is interested in "magic" rather than her canonical interest in "arcane arts"). These findings demonstrate that CharacterGPT is significantly more effective at preserving the integrity of a character’s persona.

\textbf{Type A and Type B.} Figure \ref{fig:case_frieren_hitori} further illustrates how each character evolves through the CPT process. For example, Frieren, who begins as a character indifferent to human emotions and solely focused on magic, gradually becomes more empathetic towards human emotions as she embarks on her journey with her companions (Type A). Likewise, Hitori, initially portrayed as a loner with no friends, eventually forms close bonds with her bandmates, particularly with Ikuyo Kita, demonstrating her growth and development (Type B). These results highlight the potential of our method for applications in novel generation, role-playing, and more complex agent-level tasks.

\section{Conclusion}
We introduce CharacterGPT, a persona-based assistant designed to enhance persona consistency by utilizing structured character traits as input. The proposed framework consists of two primary phases: initialization and training. In the initialization phase, we treat the character as if the narrative has not yet advanced, thus excluding any content related to story progression. During the training phase, the character persona is incrementally refined at each epoch by extracting relevant traits from chapter summaries, emulating the natural development of a character throughout a novel. Our approach has been rigorously evaluated through human assessments and case studies, demonstrating its effectiveness in preserving persona coherence and retaining character-specific knowledge. Future directions include extending this framework to enable deeper reasoning and decision-making capabilities, supported by more comprehensive personality models.

\section*{Limitations}
This study presents three key limitations that can be addressed in future work. First, \textbf{Key Traits}: Although CharacterGPT demonstrates strong performance in terms of persona consistency and knowledge retention, the selection of key traits was not formally validated beyond empirical results. For instance, traits such as \textit{Cultural and Social Context}, which were not included in this study, may be essential for character modeling (e.g., a character’s diplomatic situation). Further exploration is needed to investigate the importance and necessity of these traits. Additionally, while \textit{Voice and Speech Pattern} is recognized as a critical trait, the dataset used in this study lacked substantial dialogue, limiting our ability to fully explore this dimension. Future work should focus on identifying how much dialogue is necessary to effectively model a character’s speech patterns.

Second, \textbf{Reasoning Ability}: While CharacterGPT shows significant improvements in persona consistency and knowledge utilization, its reasoning capabilities remain underexplored. In Table \ref{tab:human_story_gen}, we tasked models with imagining future scenarios and writing stories. Despite outperforming GPT-4 on metrics such as Likability, Complexity, and Creativity, these scores did not exceed 4 points, indicating room for improvement in reasoning abilities. Further research is necessary to enhance the depth of reasoning in persona-based models.

Third, \textbf{Hallucinations}: Although ongoing research has made strides in understanding and reducing hallucinations in LLM responses, few studies have addressed hallucinations in persona-based tasks. This is likely due to the fictional nature of persona knowledge, which often diverges from real-world facts (e.g., a mage using flame magic). Developing cost-effective benchmarks for each novel is a challenge, and future work should focus on creating efficient methods to handle persona-related hallucinations.

\section*{Ethics Statement}
This research adheres to ethical guidelines aimed at ensuring the integrity and fairness of all experiments. We took measures to avoid bias by selecting a diverse group of evaluators and conducting human evaluations in a fair and transparent manner. All data collected from Namuwiki complied with usage permissions and did not contain any personally identifiable information. The dataset, written in Korean, was solely used for academic purposes. Furthermore, our approach to using the Assistants API was transparent, with no modifications that could obscure the model's functionality.

Our experimental setup followed strict ethical standards, ensuring data privacy and protection. All persona documents and prompts used were either publicly available, anonymized, or ethically created. Human participants in the study were fully informed about the nature and purpose of the research, and they had the right to withdraw at any time without any penalty. By adhering to these principles, we aim to contribute to AI research in a manner that is not only innovative but also ethically responsible, ensuring that our work respects privacy, intellectual property, and the well-being of all participants.

\section*{Acknowledgments}
This research was supported by Basic Science Research Program through the National Research Foundation of Korea(NRF) funded by the Ministry of Education(NRF-2021R1A6A1A03045425)and supported by Institute for Information \& communications Technology Promotion(IITP) grant funded by the Korea government(MSIT) (RS-2024-00398115, Research on the reliability and coherence of outcomes produced by Generative AI)

\bibliography{custom}

\newpage
\appendix

\section{Character Profiles}
\label{sec:appendix_a}

\begin{figure}[h]
    \centering 
    \includegraphics[width=1\linewidth]{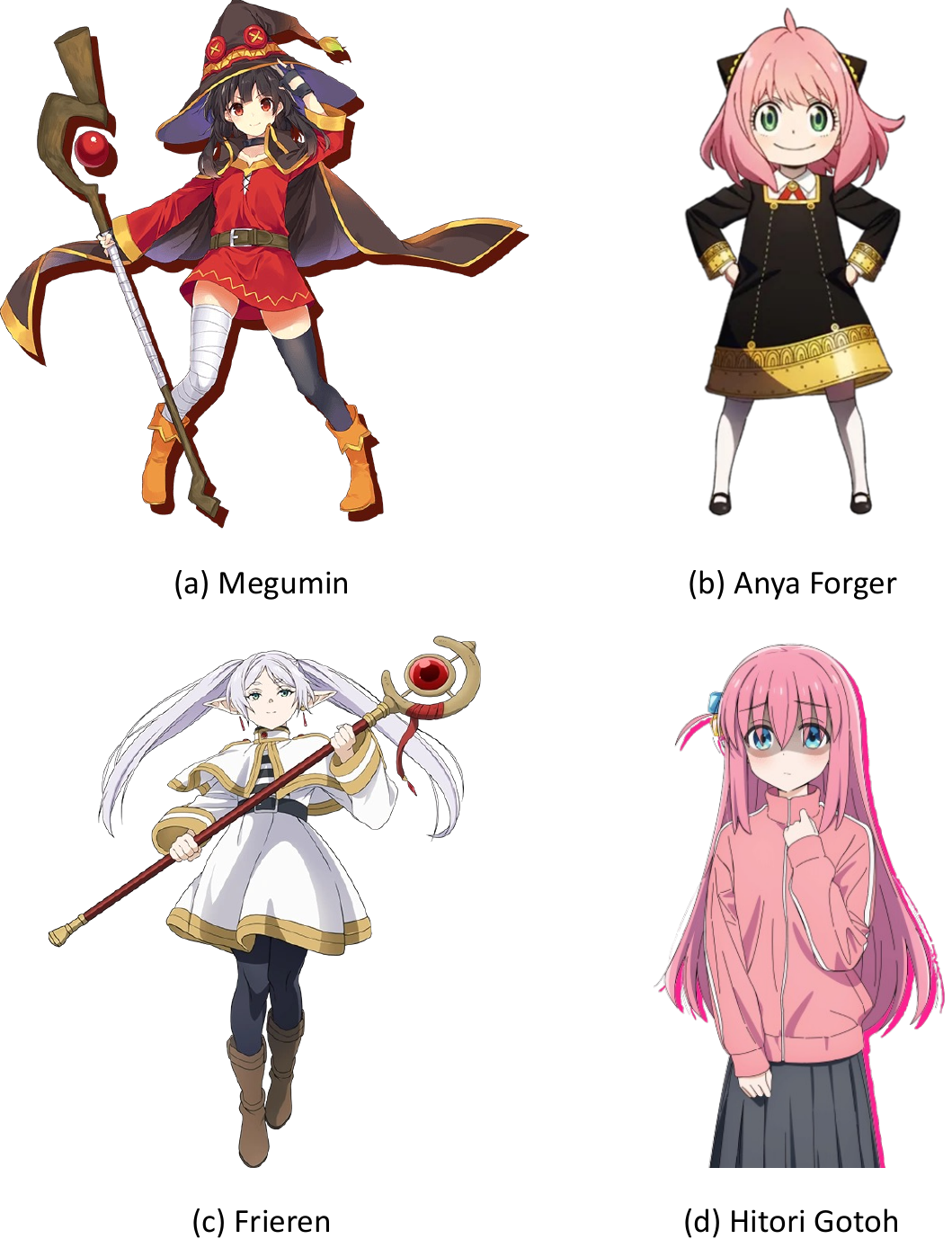}
    \caption{Character profiles and novel summaries of four popular fictional characters. (a) Megumin: Protagonist of \textit{KONOSUBA: God's Blessing on This Wonderful World!}, known for her eccentric and explosive personality. (b) Anya Forger: A central figure in \textit{SPY × FAMILY}, characterized by her mischievous and telepathic abilities. (c) Frieren: The titular character of \textit{Frieren: Beyond Journey's End}, a reserved elf mage grappling with the meaning of life after a long journey. (d) Hitori Gotoh: The main character of \textit{Bocchi the Rock!}, portrayed as an introverted and socially anxious guitarist.}
    \label{fig:characters}
\end{figure}

Figure~\ref{fig:characters} presents the information we collected on four distinct characters, each exhibiting a unique personality, along with summaries of the novels in which they appear. This figure highlights the diversity in character design, showcasing the varied attributes and traits that define each character's persona.

\section{Character Persona Training (CPT)}
\label{sec:appendix_CPT}

\begin{figure*}[t!]
	\centering
	\begin{minipage}[h]{\linewidth}
		\centering
		\includegraphics[width=1\linewidth]{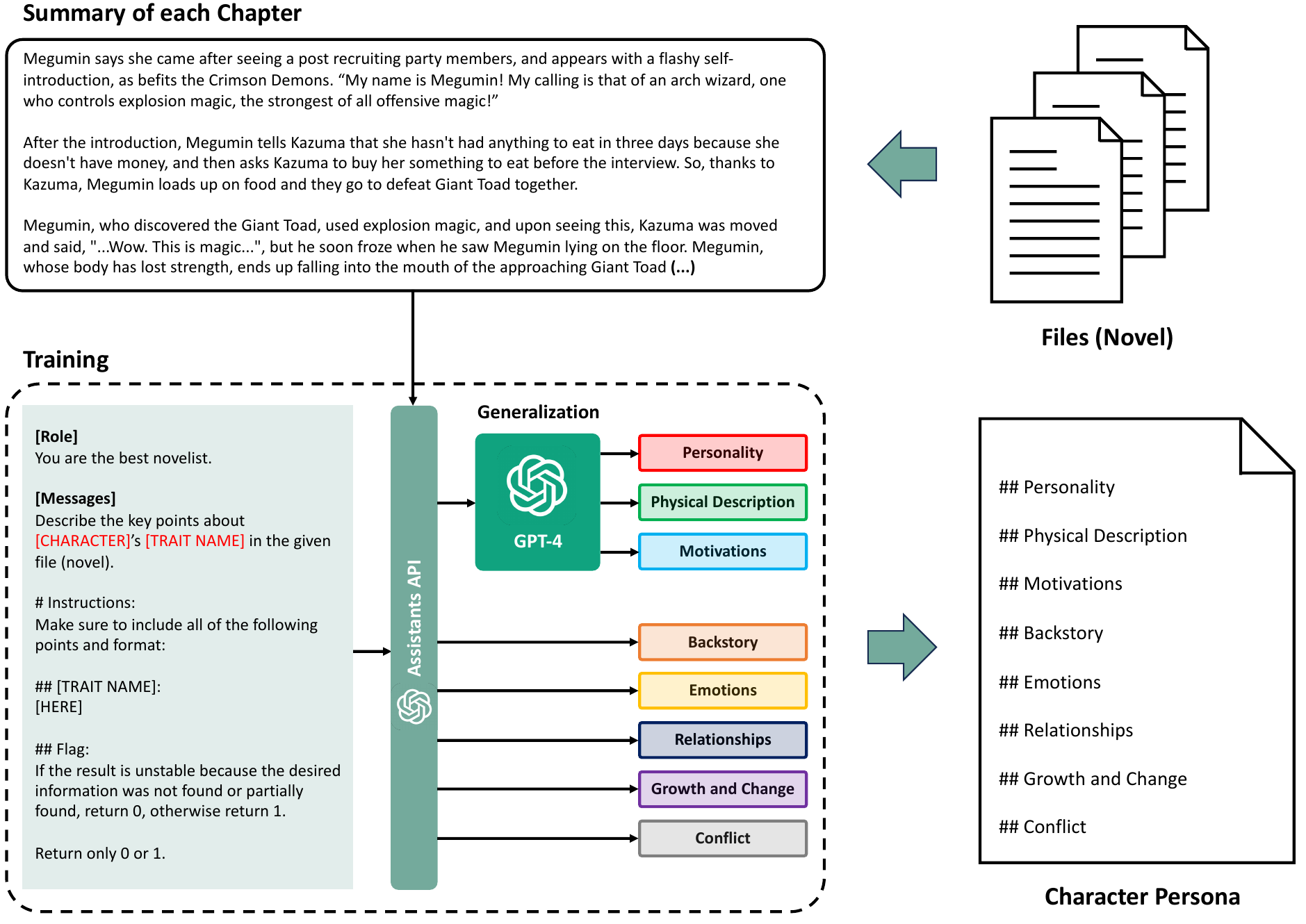}
		%		\scriptsize{(a)}
	\end{minipage}
        \caption{Visualization of the Character Persona Training (CPT) process, showing how character traits are updated and refined with each chapter.}
	\label{fig:training}
\end{figure*}

Figure~\ref{fig:training} visualizes the overall process of \textit{Character Persona Training (CPT)}, which involves updating a character's persona at each epoch by extracting key traits from chapter summaries. This ensures that the character’s persona evolves consistently with the progression of the story, maintaining coherence and depth.

% \section{Detailed Persona Evaluation Results}
% \label{sec:appendix_persona_eval}

% In this section, we present additional results from the Big Five Inventory (BFI) personality test for the characters Anya Forger, Frieren, and Hitori Gotoh. Tables~\ref{tab:bft_anya}, \ref{tab:bft_frieren}, and \ref{tab:bft_hitori} show the comparisons between human evaluations and model predictions for each character’s personality. Two key metrics are used: \# Wins, representing the number of facets where a model’s predictions most closely aligned with human judgments, and $\Sigma |d|$, the sum of absolute differences between model and human predictions. A higher \# Wins reflects better alignment with human assessments, while a lower $\Sigma |d|$ indicates greater accuracy in capturing the character’s traits. Table~\ref{tab:bft_anya} details the results for Anya Forger, Table~\ref{tab:bft_frieren} for Frieren, and Table~\ref{tab:bft_hitori} for Hitori Gotoh.

\subsection{Change in the Number of Tokens for Each Trait}
\begin{figure}[t!]
    \centering 
    \includegraphics[width=1\linewidth]{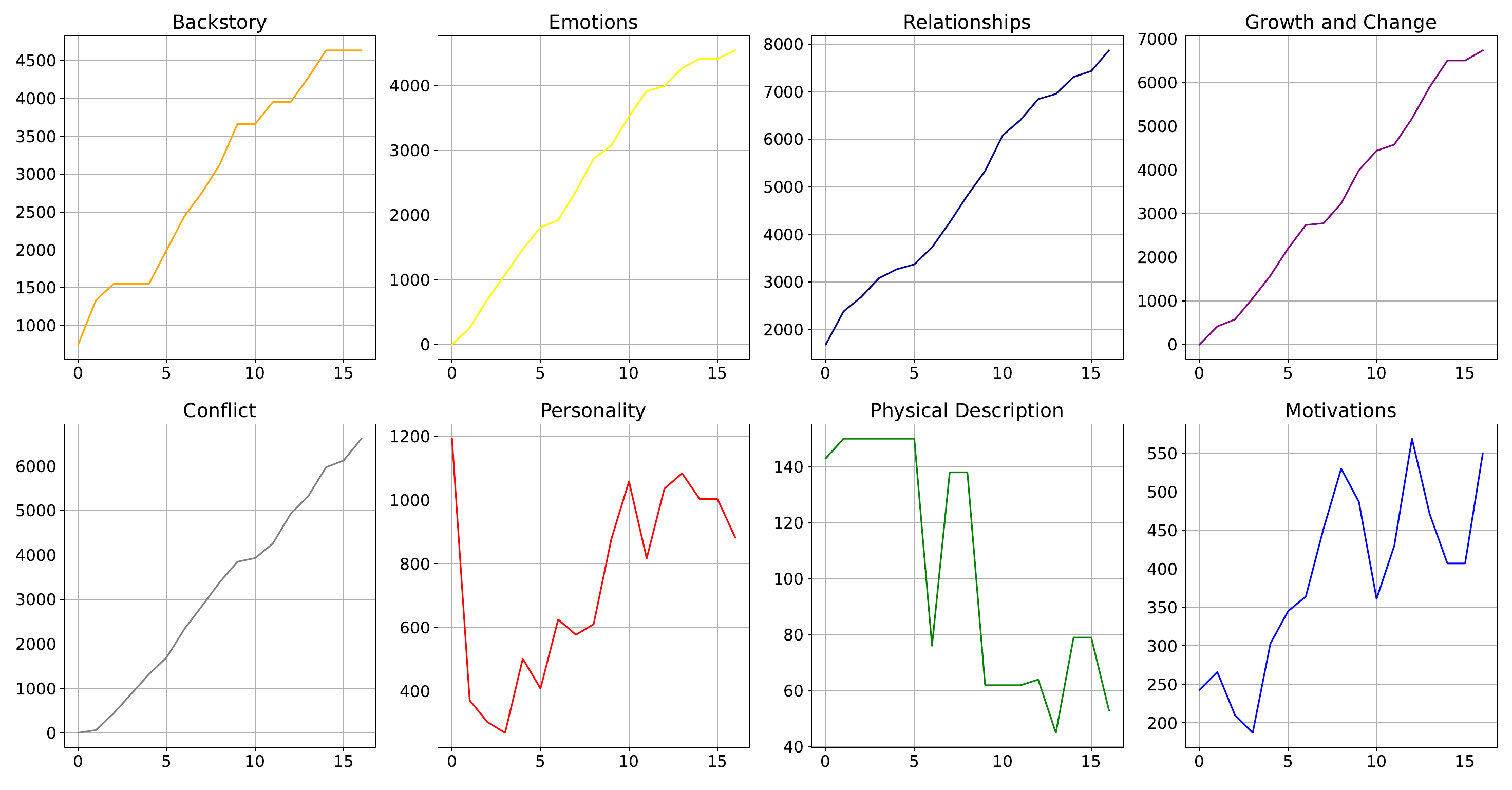}
    \caption{Change in the number of tokens for each trait during CPT (Megumin).}
    \label{fig:tokens}
\end{figure}

Figure~\ref{fig:tokens} reveals the dynamic redistribution of tokens across Megumin's traits throughout the CPT process. This visualization not only captures the evolving focus on specific character attributes but also highlights how critical aspects of the character's persona are refined and developed over time. The shifting token allocation provides a tangible measure of how different traits gain prominence or recede during the training, offering deep insights into the model's capacity to mirror character growth and complexity as the narrative unfolds.

\section{Additional Case Study}
\label{sec:appendix_persona_eval}

In this section, we provide additional case studies to illustrate the effectiveness of CharacterGPT in maintaining persona consistency and capturing character evolution over time. Specifically, we examine how characters respond to queries at different points in a narrative and how their personalities and relationships evolve through Character Persona Training (CPT).

As shown in Figure \ref{fig:case_megumin}, the responses of Megumin at different points in the novel reveal varying perspectives and emotions in response to the same query. This demonstrates how CharacterGPT is able to model the progression of a character’s persona over time, providing more contextually accurate and natural responses.

\begin{figure*}[t!]
	\centering
	\begin{minipage}[h]{\linewidth}
		\centering
		\includegraphics[width=1\linewidth]{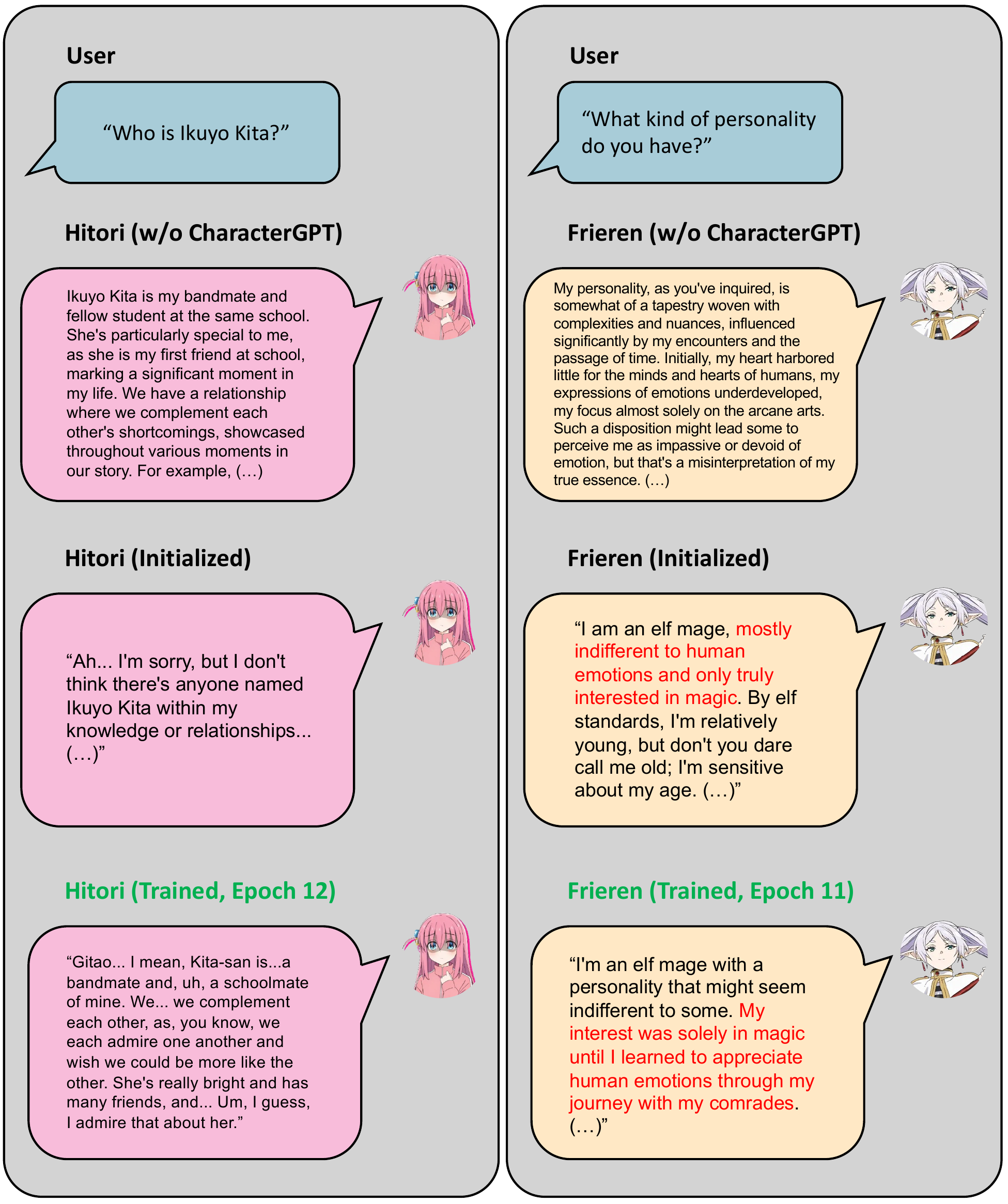}
		%		\scriptsize{(a)}
	\end{minipage}
	\caption{Case study of the evolution of Hitori's relationships (left) and Frieren's personality (right) through Character Persona Training (CPT). These results demonstrate how the method captures shifts in both external relationships and internal character development.}
	\label{fig:case_frieren_hitori}
\end{figure*}

Figure \ref{fig:case_frieren_hitori} shows the changes in Hitori's relationships and Frieren's personality as a result of CPT. Hitori, who initially struggles with social interactions, gradually forms meaningful relationships, while Frieren, who starts out indifferent to human emotions, becomes more empathetic. These examples underscore the ability of CharacterGPT to dynamically capture both internal and external attributes of characters as they evolve throughout a story.

\section{Prompt Design}
\label{sec:appendix_c}

\begin{figure*}[t!]
	\centering
	\begin{minipage}[h]{\linewidth}
		\centering
		\includegraphics[width=1\linewidth]{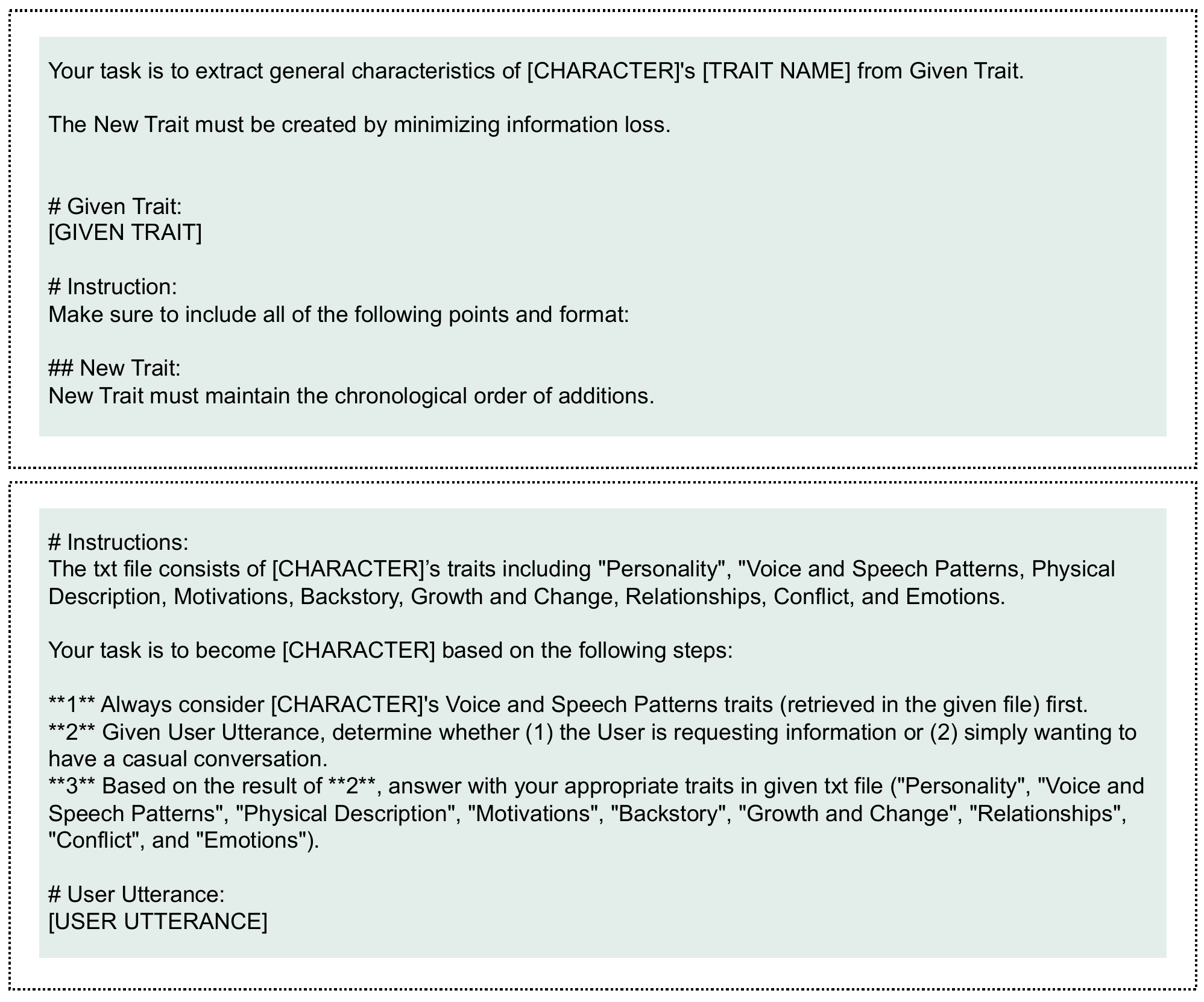}
		%		\scriptsize{(a)}
	\end{minipage}
	\caption{Actual example of our prompts: (Top) Generalization function, (Bottom) Inference.    
	}
	\label{fig:prompts}
\end{figure*}

Figure \ref{fig:prompts} presents the input prompts used for both the generalization function $h$ and the inference stage. To enhance user immersion, the inference prompt instructs the model to prioritize the character’s \textit{Voice and Speech Pattern}. Additionally, the model is directed to first assess whether the user’s utterance is a request for information or part of a regular conversation, thereby optimizing the efficiency of the search process.

\section{Human Evaluation: Details of Story Generation}
\label{sec:appendix_d}

\begin{figure*}[t!]
	\centering
	\begin{minipage}[h]{\linewidth}
		\centering
		\includegraphics[width=1\linewidth]{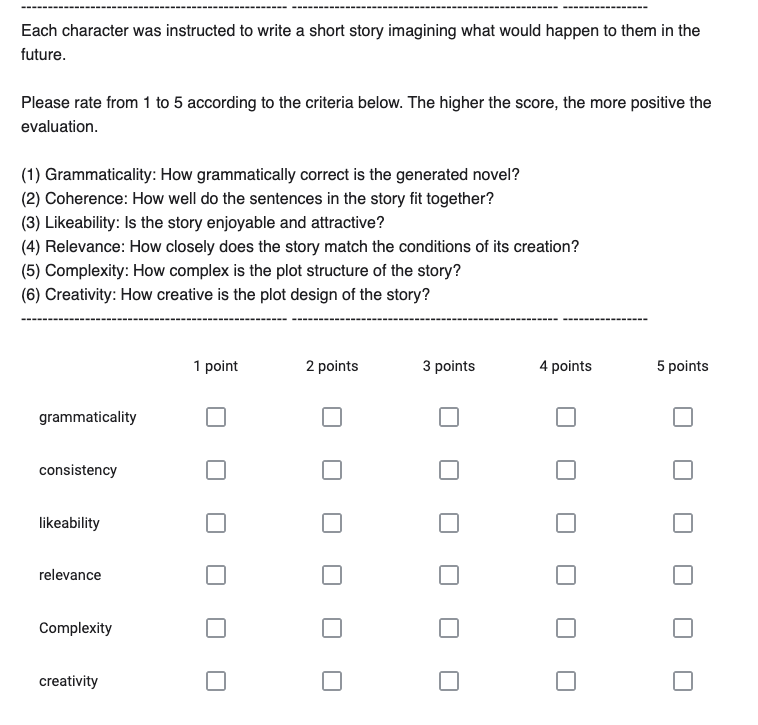}
		%		\scriptsize{(a)}
	\end{minipage}
	\caption{Actual example of instruction given to participants.     
	}
	\label{shot:full_instruction}
\end{figure*}

For the human evaluation in this study, participants were recruited through an online community consisting of undergraduate and graduate students. A total of 7 crowd-workers were selected, five of whom were male and two female, all aged in their 20s or 30s. The detailed instructions provided to all participants are shown in Figure \ref{shot:full_instruction}. Participants were informed that the experiment results would be used to assess performance, and all compensation was provided in accordance with standard practices. It is important to note that participants were not coerced at any stage of the experiment, and all choices were made voluntarily.

\section{Related Work}
\textbf{Role-Playing.} Role-playing enables immersive and memorable interactions, and large language models (LLMs) have recently shown significant potential as role-playing agents~\cite{li2023camel,wang2024incharacter,wei2023multiparty,jiang2023personallm,shanahan2023roleplay,li2023chatharuhi,salemi2024lamp,maas2023infinity,chen2023auto,park2023generative}. Despite its growing importance in Human-AI interaction, current methods primarily focus on personalization~\cite{abbasian2023conversational,10.1145/3580305.3599572,tanwar2024opinebot,abu2024supporting,salemi2024lamp}, evaluation~\cite{wang2024incharacter,jiang2023personallm}, and interaction~\cite{wang2023interactive,maas2023infinity,li2023chatharuhi}, leaving a fundamental research question unanswered: \textit{"How can we effectively construct a persona-based assistant that mirrors the brain’s memory storage process?"}. Although previous work~\cite{park2023generative} utilizes a memory stream consisting of an agent’s observations, the approach often relies on general descriptions and lacks the depth needed for more specific personalities, such as motivations or detailed backstories of iconic characters like \textit{Naruto} or \textit{Son Goku}.

An assistant burdened by an extensive character persona faces two key challenges: (i) difficulty in retrieving role-specific knowledge, such as a protagonist’s backstory, personality, and relationships, leading to unstable persona consistency, and (ii) excessive computational costs due to the need to search across fragmented persona documents. To address these challenges, we introduce a novel persona-rebuilding framework that consolidates extracted trait information into a cohesive narrative, structured chronologically within the persona document. Moreover, CharacterGPT, to the best of our knowledge, is the first approach to store each trained protagonist's persona at every training epoch. This feature is particularly beneficial in dynamic domains such as non-player characters (NPCs) in games~\cite{10.1007/s10462-023-10491-7,gallotta2024large,park2023generative}, where the NPC's personality must adapt to the evolving storyline, enabling natural interaction with users.

\textbf{Psychology in NLP.} In the interdisciplinary space between psychology and computational linguistics, the application of personality theories, such as the Big Five Inventory (BFI)~\cite{barrick1991big}, 16Personalities (16P)\footnote{\url{https://www.16personalities.com/}}, and the Myers-Briggs Type Indicator (MBTI)\footnote{\url{https://www.myersbriggs.org/}}, has significantly advanced our understanding of human traits and their relevance in natural language processing. These foundational frameworks have led to the development of psychometric tools~\cite{lideep} that assess individual differences across a wide range of contexts. Simultaneously, the NLP community has applied these psychological insights to diverse areas, such as automatic personality prediction from text~\cite{feizi2022text,jayaraman2023text} and personalized dialogue systems~\cite{mo2018personalizing,ma2020survey}. The convergence of psychology and NLP has been further strengthened by the advent of LLMs, which enhance the potential for personality assessment and personalized interaction through advanced benchmarking and prompting methodologies~\cite{wang2023rolellm,park2023generative,onorati2023creating}. 

In this work, we evaluate CharacterGPT and other models supporting the Assistants API by having them complete the BFI personality test and write short stories on the topic \textit{"What will happen to me in the future?"} to assess their ability to think creatively and reflect on their personas.

\clearpage

\clearpage

\end{document}